\documentclass[conference]{IEEEtran}
\usepackage[colorlinks=true, linkcolor=black, citecolor=black, urlcolor=black]{hyperref}
\usepackage[table]{xcolor} 
\usepackage{array}       
\usepackage{tabularx}
\definecolor{tableheader}{gray}{0.93} 
\usepackage{xurl} 
\usepackage{float}
\IEEEoverridecommandlockouts
\usepackage{cite}
\usepackage{amsmath,amssymb,amsfonts}

\usepackage{booktabs}
\usepackage{tabularx}
\usepackage{array}

\usepackage{algorithmic}
\usepackage{graphicx}
\usepackage{subcaption}
\usepackage{textcomp}
\usepackage{xcolor}
\def\BibTeX{{\rm B\kern-.05em{\sc i\kern-.025em b}\kern-.08em
    T\kern-.1667em\lower.7ex\hbox{E}\kern-.125emX}}
\begin{document}

\title{Social Engineering Attack Detection in Corporate Communications \\
    \thanks{The code for this paper can be found in this link: \url{https://github.com/jhapappu7165/Social_Engineering_Attack_Detection_In_Corporate_Communications}.}
}

\author{
    \IEEEauthorblockN{Hanzla Hamid, Pappu Jha, Nick Rahimi}
    \IEEEauthorblockA{
        University of Southern Mississippi, Hattiesburg, Mississippi, USA\\
        \texttt{\{hanzla.hamid, pappu.jha, nick.rahimi\}@usm.edu}
    }
}

\maketitle

\begin{abstract}

Social Engineering attacks continue to pose a significant risk to the corporate world with a growing level of sophistication. It is difficult to distinguish between authentic and phishing communication via email, chat, and phone. They pose a risk of exposing trade secrets and confidential communications and can create a chain of attack from lower-level employees up to executives. Therefore, we study this growing threat using real-world datasets to build baseline machine-learning models. Our TF-IDF models classify emails as phishing or safe, achieving accuracy above 90\%. We build on this by exploring whether psychological manipulation and persuasion indicators can be measured to play a significant role in detecting phishing. Instead of relying solely on surface-level lexical features, we derive interpretable lexicon-based indicators of urgency, trust, greed, and other persuasion cues from the combined email dataset. By evaluating these manipulation-oriented features on their own, we measure how much signal they carry (ROC-AUC of 0.74 on a held-out test split) and test whether they transfer to held-out email sources.

\end{abstract}

\begin{IEEEkeywords}
social engineering, machine learning, communication
\end{IEEEkeywords}

\section{Introduction}
Social engineering is a form of cyber attack that manipulates people into sharing information, downloading software, visiting websites, or sending money that can compromise their personal, financial, or organizational security \cite{ibm_social_engineering}. In most cases, threat actors adopt some of the commonly used tactics, including posing as a trusted brand or a government agency and appealing to greed or helpfulness.

Social engineering attacks are executed in several forms, namely phishing, baiting, tailgating, pretexting, and scareware. To summarize each of them, phishing is the most popular social engineering attack, which involves manipulating individuals into performing actions like clicking on malicious links or opening infected files that can harm their device or their organization's network. Baiting, on the other hand, includes tempting a victim with a valuable offer to concede to the demands of a bad actor. Tailgating, meanwhile, means unauthorized access by a threat actor due to a physical or digital security breach. Pretexting takes a different approach, designing a false scenario and making the victim believe that the threat actor is the right person to solve it. Scareware, finally, refers to utilizing digital services to create fear among people and manipulate them into performing dangerous activities.

In the corporate world, social engineering attacks have been a major threat to the proprietary data, financial health, and reputation of organizations. Some of the high-profile incidents in recent years include the theft of around \$400k from a Shark Tank judge \cite{nbc2020_corcoran} and the loss of about \$37 million by a Toyota Boshoku subsidiary \cite{bleeping2019_toyota}. All such events resulted from the wrongful influence on employees. Likewise, this research was also inspired by a real-world observation of how a social engineering attack involving an email addressed to an intern using the manager's identity was narrowly avoided at a major company in Mississippi.

Therefore, it is essential to have a robust detection system dedicated to the corporate world that can classify an incoming email into phishing and safe categories in real-time, \textbf{and also analyze the psychological manipulation attempts.} Notably, email communication was used for the study because it is the preferred mode of communication in the current professional world.

Our contributions to this paper include the following:
\begin{itemize}
\item We develop and evaluate multiple machine learning models for detecting social engineering attacks in email communications, establishing baseline performance and comparing the effectiveness of different classification approaches.

\item We investigate the role of psychological and emotional characteristics in social engineering detection by incorporating emotion- and intent-based linguistic features derived from email text.

\item We construct a diverse, deduplicated dataset (82{,}078 emails) from six email collections representing different contexts, including publicly available organizational email exchanges, to provide a broader basis for evaluating social engineering detection.

\item We apply statistical analysis, including the Mann--Whitney U test, to examine whether differences in feature distributions between malicious and benign emails are statistically significant rather than attributable to chance.

\item We investigate potential dataset-specific distributional artifacts by identifying and removing features that may introduce information leakage, and evaluate a leakage-resistant feature set, including on held-out email sources, to assess whether psychological and intent-based features retain predictive value.

\item Our results demonstrate that psychological and intent-based linguistic features provide meaningful predictive signal for social engineering detection, with the leakage-resistant model achieving a ROC-AUC of 0.7397 on unseen test data.

\end{itemize}

This paper is structured as follows: the next section (II) provides a detailed account of existing studies conducted in the context of social engineering attacks and the role of machine learning involved. Similarly, Section III explains the methodology, including data collection, preprocessing, feature extraction, model training, testing, and evaluation. Likewise, Section IV presents the results and discusses the performance of the models. Section V describes the limitations of the study. Finally, the last section (VI) concludes the study with notable findings and makes recommendations for future studies.

\section{Literature Review}
Social engineering attacks are a unique type of threat in that they mainly take advantage of human psychology and error rather than technical vulnerabilities. This means that it is generally harder to defend against such attacks through traditional security measures. Mashtalyar et al. provide in their paper a general review of the modern threat environment and argue that social engineers aim to exploit the psychological weaknesses of individuals through various methods that weaponize manipulation and deception \cite{mashtalyar2021_social_engineering}. The complexity of such attacks has evolved with the advancement of the digital infrastructure that organizations rely on.

Supporting this argument is the Delphi study conducted by Campbell, which involved twenty information security experts \cite{campbell2018_deception}. This study found that the three major organizational issues --- compromised data, ineffective practices, and inadequate ongoing education --- revolve around one core cause: insufficient understanding of how human deception operates in the context of corporate communications. 

Detecting attacks first requires that a range of classifications of attack types and techniques be established. For this purpose, Odeh et al. survey the primary categories of social engineering attacks with the use of existing threat detection and prevention tools \cite{odeh2021_survey}. They use these categories to record and classify the psychological triggers that are most utilized by attackers, ranging from urgency and authority to scarcity and reciprocity. The main gap shown by their study is that even though rule- and keyword-driven filters are widely employed, they prove to be weak against context in language, which has driven the use of lexicon-based feature engineering that is rooted in intent in this research, rather than merely utilizing surface-level terms.

Due to the inadequacy of rule-based systems in accurately generalizing over the diversity of language in realistic attacks, machine learning-based applications are becoming more and more prevalent. This is directly addressed by Wang et al., as he re-contextualizes social engineering threat detection as an ML problem rather than one solely defined by an attack vector, such as email phishing \cite{wang2022_ml_social_engineering}. Their work has direct methodological relevance to this study because, like Wang et al., this research converts message content into structured and interpretable feature vectors rather than treating email classification as merely a raw text problem.

Ogunlade assesses the role of AI in modern phishing detection, evaluating techniques such as Natural Language Processing (NLP) and multimodal detection to identify threats, including phishing and deepfake impersonation \cite{ogunlade2025_ai_social_engineering}. Their study highlights the declining effectiveness of traditional detection methods against increasingly sophisticated social engineering attacks. Ogunlade suggests that NLP methods focusing on the persuasive intent of messages, rather than only their technical or structural features, may prove to be a stronger approach to threat detection. This supports the implementation used in the present study, where persuasive language patterns such as excitement, trust, curiosity, urgency, and neutral formality are converted into measurable features. These features are then evaluated using logistic regression and Mann-Whitney U statistical testing on a combined email dataset to determine their usefulness in distinguishing phishing emails from legitimate messages.

A further line of work studies persuasion principles in phishing directly. Ferreira et al. merged the principles of Cialdini, Gragg, and Stajano et al. and showed, through manual analysis of phishing emails, that these principles appear in specific combinations \cite{ferreira2015_principles}. Wright et al. found in a field experiment that different phishing influence techniques differ in how dangerous they are \cite{wright2014_influence}, and a recent survey by Khadka et al. identifies a significant gap in understanding how persuasion principles affect phishing outcomes \cite{khadka2024_persuasion_survey}. These studies motivate our lexicon categories, but they rely on manual analysis or user experiments, whereas we operationalize the cues as automatic lexicon features and test them at scale.

Overall, these studies provide the conceptual and methodological foundation for the present research. Previous literature shows that social engineering attacks are largely successful due to the fact that they exploit human psychology \cite{mashtalyar2021_social_engineering,campbell2018_deception}. Research on psychological triggers such as urgency, authority, trust, and curiosity has also shown their importance in both creating and detecting these attacks \cite{odeh2021_survey}. Machine learning approaches using structured and interpretable features extracted from message content have also been shown to be practical and effective for detecting social engineering threats \cite{wang2022_ml_social_engineering,ogunlade2025_ai_social_engineering}. However, existing research has not fully explored how emotional and intent-based language patterns differ between malicious and legitimate corporate emails through statistical validation using interpretable models. This study addresses this gap by analyzing these patterns and evaluating their effectiveness for phishing detection.

\section{Methodology}

As the study is based on the detection of social engineering attacks in corporate emails, we focused on a public dataset representing such real-world communication. The dataset was available on Kaggle with the title "Phishing Email Dataset" by Naser Abdullah Alam. The work commenced with training baseline models to categorize an email into a malicious or benign category. Later, the focus shifted to identifying and ranking the intent and emotions of the emails based on Cialdini's Principles of Persuasion. Finally, a second set of classification models was trained using only these intent and emotion features, with no access to the raw email text, to measure how much predictive signal the psychological indicators carry on their own.

\subsection{Email Classification}

To classify incoming email messages into proper categories, the following processes were involved: (1) Data collection, (2) Exploratory Data Analysis, (3) Feature Engineering, (4) Model
Training, and (5) Analysis.

Beginning with the first step, we collected 7 datasets from Kaggle. They include CEAS\_08.csv, Enron.csv, Ling.csv, Nazario.csv, Nigerian\_Fraud.csv, SpamAssassin.csv, and phishing\_email.csv. It is worth noting that Enron.csv and Ling.csv contain information on the core content of phishing emails, including columns such as subject, body, and labels, which indicate whether an email is phishing or safe. On the other hand, CEAS\_08.csv, Nazario.csv, Nigerian\_Fraud.csv, and SpamAssassin.csv provide more detailed information about the emails, incorporating sender and receiver details, date, and classification labels. 

Since phishing\_email.csv already aggregates the other six files, it was used as the combined dataset; using all seven files would duplicate emails. Firstly, the subject and body columns of emails were combined to generate one input column. Likewise, any email records with null values, i.e., empty records in the text and label columns, were dropped from the datasets. Duplicate emails were then removed on the combined text, before any splitting or cross-validation, so that identical emails cannot fall in both training and test folds; this removed 408 emails and left 82{,}078 unique ones. Similarly, a TF-IDF vectorizer was applied to the texts by breaking them into words and then turning them into numeric features (re-fit on the training folds only) before using them for training. To understand the size of the dataset, several analyses were done, such as finding the number of email texts in each of the two categories. During the training of the model, a 5-fold cross-validation was performed, with four folds used for training and the remaining one for testing, and the process was repeated five times. In order to gain insights into how the model performs with varying sizes of training and validation data, a learning curve was drawn. Not to mention, the gap between the training and validation curves represents overfitting. Finally, the ROC-AUC and Precision-Recall curves were drawn to further understand the model's ability.

For categorizing emails into safe or phishing categories, we trained two baseline models using Logistic Regression and Naive Bayes algorithms.

\subsection{Psychological Manipulation}

This section details the end-to-end pipeline designed to detect social engineering attacks through the lens of analyzing emotion and intent. The three main principles that guide this workflow are interpretability, leakage resistance, and statistical rigor. Instead of treating email attack detection solely as a text processing problem, this method converts each email into a structured numerical vector representing its communicative intent and emotional tone. A logistic regression classifier is then applied, and its coefficients can be directly inspected and explained.

\subsubsection{Dataset and Preprocessing}

This part of the research used the same deduplicated combined dataset as the baseline classifiers (82{,}078 emails), not CEAS\_08 alone. Each record consists of the combined subject-and-body text and a binary label which indicates whether the email is malicious (1) or benign (0). The combined dataset was used because it mixes corporate, personal, spam, and fraud email from several sources, which allows us to test whether the features hold up beyond a single collection.

The raw email body text field was first cleaned before feature extraction. This was done to reduce noise and also ensure consistency. The preprocessing flow consisted of four main stages. First, HTML tags and markup were removed. Next, URLs and email address strings were stripped. All of the email body text was then converted to lowercase. The preprocessing pipeline was concluded by separating the text into individual word tokens through whitespace-normalized tokenization. This process was vital as it ensured that lexicon matching was done on normalized meaningful terms rather than artifacts of email formatting or sender metadata. 

\subsubsection{Feature Engineering}
 
Rather than working directly on raw text, each email was first transformed into a feature vector based on predefined lexicon categories, each representing a specific type of intent or emotion: urgency, trust, authority, excitement, fear, greed, curiosity, social\_connection, request\_action, promotion\_marketing, information\_transactional, sexual\_pleasure, technical\_discussion, and neutral\_formal. The choice to work at this level, rather than with raw word counts or embeddings, follows from the theoretical viewpoint established in the literature \cite{ibm_social_engineering, mashtalyar2021_social_engineering, campbell2018_deception}: social engineering succeeds by exploiting specific psychological triggers, so detecting those triggers directly at the lexical level should provide a detection signal that is both useful and, just as importantly, explainable.
 
Several of the categories were chosen with Cialdini's principles of persuasion in mind \cite{cialdini2021_influence}, a framework widely used in the social engineering literature to describe the compliance mechanisms attackers rely on. The \textit{authority} category tracks Cialdini's authority principle directly: messages that invoke a position of power, a supervisor, or an official institution tend to be obeyed with less scrutiny. \textit{Urgency} corresponds to the scarcity principle, where a manufactured, limited time window pushes the recipient to act before they have a chance to think it through. \textit{Social\_connection} draws on social proof and liking, capturing language that invokes shared relationships, familiarity, or group belonging. \textit{Trust} sits closer to liking and reciprocity, picking up attempts to build credibility or rapport before a request is made. \textit{Request\_action} maps onto the consistency principle, which plays on a recipient's tendency to follow through once a small, reasonable-seeming commitment has already been framed. Not every category fits neatly into Cialdini's six principles, though: greed, excitement, curiosity, and fear are grounded instead in the broader emotion-based persuasion literature on social engineering \cite{ibm_social_engineering, mashtalyar2021_social_engineering, odeh2021_survey, ogunlade2025_ai_social_engineering, mithun2024_phishing_ai, duan2025_emotional_lens, tshimula2024_psych_profiling, toth2025_phishing_dataset}, which treats emotional arousal and incentive-based appeals as additional levers attackers reach for alongside, and often together with, compliance-based tactics. Grounding the lexicon this way was meant to tie the feature engineering back to an established theoretical basis rather than leaving the category list to read as an arbitrary set of keyword groupings.
 
Each category was built from a hand-curated set of seed terms and short phrases meant to represent its underlying psychological construct. The \textit{urgency} category, for instance, includes terms such as ``immediately,'' ``act now,'' and ``expires today,'' while \textit{authority} includes phrases like ``your manager,'' ``IT department,'' and ``official notice.'' These term lists, and the category boundaries themselves, were developed iteratively by going through known phishing and legitimate email samples by hand, rather than by validating against an existing third-party lexicon. The categories with an explicit grounding in Cialdini's framework rest on somewhat firmer theoretical footing than the purely emotion-based ones.
 
For each email and each lexicon category, three statistics were calculated: the raw count (how many tokens in the email matched the category), binary presence (whether any matching token appeared at all), and normalized frequency (the raw count divided by the email's total token count). We settled on normalized frequency for the final models, since it accounts for variation in email length and lets features be compared directly across emails of very different sizes.
 
The result is that each email is represented as a vector where every element is a normalized score for one lexicon category, which keeps the whole feature space interpretable: each number simply says what fraction of an email's content falls into a given emotion or intent category. An email with high urgency and trust scores, for example, reads as one that is applying time pressure while also trying to establish credibility, a combination consistent with the phishing strategies discussed earlier \cite{ibm_social_engineering,campbell2018_deception}.
 
\subsubsection{Modeling}
 
Logistic regression was chosen as the model for this classification task for two main reasons. First, and most importantly, it is directly interpretable: each feature's coefficient reflects its contribution, in log-odds, to the probability of a malicious label, so the model's reasoning stays transparent rather than hidden inside a black box. Second, it offers a stable baseline for a linearly separable feature space, and its ROC-AUC score holds up as a reliable performance measure even when the underlying dataset is imbalanced.
 
Two model configurations were evaluated. Model 1 used the full set of extracted lexicon features. On inspecting the fitted coefficients, three categories stood out, technical\_discussion, sexual\_pleasure, and promotion\_marketing, each with unusually large coefficients. That pattern pointed to the model picking up dataset-specific quirks rather than a signal that would generalize. The technical\_discussion category, for example, was almost entirely confined to legitimate emails, which made it a strong predictor within this particular dataset but not one we could trust as a genuine marker of intent. Model 2 removed these three leakage-prone categories and was retrained on what remained; it, in its standardized form, is the model we treat as primary throughout this study, since it gives a more honest estimate of how useful intent- and emotion-based features actually are for detecting social engineering.
 
After deduplication, the dataset used for this part of the study came to 82{,}078 unique emails, 42{,}845 malicious and 39{,}233 legitimate, drawn together from several established phishing and legitimate email sources. We wanted the scale and variety here on purpose: training on a broad, heterogeneous mix of email styles, rather than one smaller source, gives a more honest sense of whether intent- and emotion-based features actually hold up, as opposed to just fitting the quirks of a single collection.
 
One wrinkle is worth flagging here. L2-regularized logistic regression doesn't penalize every feature equally when they sit on very different raw scales, and our normalized-frequency features vary a fair amount in variance from one category to the next. To get around this, we also fit a standardized version of Model 2, rescaling every feature to zero mean and unit variance (the scaler was fit on the training partition only, so no test-set information leaks in) before training. In an unregularized model, this rescaling wouldn't touch the decision boundary at all; under L2 regularization, though, it does shift which features the penalty leans on. We therefore treat the standardized version as the main result, reporting its coefficients alongside their odds ratios ($e^{\beta}$), which express each feature's effect per standard deviation and make it easier to compare feature importance on equal footing; the unstandardized version is kept for reference.

Models were first evaluated on a stratified 80/20 train/test split (16{,}416 test emails; the malicious class is the positive class for precision, recall, and the confusion matrix). We then ran 5-fold stratified cross-validation on the full deduplicated set, with the scaler fit inside each training fold. To test transfer across sources, we also ran a leave-one-source-out evaluation: the six source files were cleaned with the same pipeline and deduplicated across sources (74{,}232 unique emails, fewer than above because empty texts are dropped and deduplication is applied to the cleaned text), each source was held out in turn, and the models were trained on the other five.
 
\subsubsection{Statistical Validation}
We also wanted some assurance that the differences we were seeing between malicious and benign emails reflected real underlying patterns and not just noise, so each feature was tested independently with a Mann-Whitney U test. This test was chosen specifically because it compares two independent groups without assuming a normal distribution, which suits email-derived features well given how skewed they tend to be and how often they sit at zero. Alongside the p-value, we also computed two effect measures on the training partition for each category: the difference in means between the two classes, and Cliff's delta, a rank-based effect size that, like the test itself, assumes no particular distribution. One caveat is worth stating up front: given the size of the dataset, nearly every feature reaches conventional statistical significance, so p-values alone stop being very informative. We lean on effect size instead as the main basis for judging a feature's practical importance, and return to this point directly in Section~\ref{sec:urgency-case}.

\section{Results and Discussion}

\subsection{Email Classification Results}

\begin{table}[htbp]
\centering
\renewcommand{\arraystretch}{1.2}
\setlength{\tabcolsep}{6pt}

\begin{tabular}{|l|c|c|c|c|}
\hline
\textbf{Model} & \textbf{Accuracy} & \textbf{Precision} & \textbf{Recall} & \textbf{F-1} \\
\hline
Logistic Regression & 0.9853 & 0.9828 & 0.9891 & 0.9860 \\
\hline
Naive Bayes & 0.9779 & 0.9851 & 0.9723 & 0.9787 \\
\hline
\end{tabular}

\caption{Comparison of the classification models (deduplicated data).}
\label{tab:model_performance}

\end{table}

All models achieved performance metrics of over 90\% while classifying emails into one of two categories. To confirm that they were not overfitting, 5-fold cross-validation was performed, and the results are presented below:

\begin{table}[htbp]
\centering
\renewcommand{\arraystretch}{1.15}
\setlength{\tabcolsep}{6pt}

\begin{tabular}{|l|c|c|c|c|c|}
\hline
\textbf{Model} & \textbf{F1} & \textbf{F2} & \textbf{F3} & \textbf{F4} & \textbf{F5} \\
\hline
Logistic Regression & 0.9886 & 0.9854 & 0.9854 & 0.9873 & 0.9876 \\
\hline
Naive Bayes & 0.9786 & 0.9776 & 0.9759 & 0.9793 & 0.9807 \\
\hline
\end{tabular}

\caption{Five-fold cross-validation F1 scores (deduplicated data).}
\label{tab:cross_validation_scores}

\end{table}

\subsection{Logistic Regression Results}
 
\begin{table}[htbp]
    \centering
    \caption{Model 1 (Full Feature Set)}
    \renewcommand{\arraystretch}{1.3}
    \begin{tabularx}{\linewidth}{| l | X |}
        \hline
        \rowcolor{tableheader}
        \textbf{Metric} & \textbf{Value} \\
        \hline
        Accuracy & 0.7382 \\
        \hline
        ROC-AUC & 0.8210 \\
        \hline
        Top malicious predictors & sexual\_pleasure (+19.37), trust (+19.00), greed (+18.75) \\
        \hline
        Top legitimate predictors & technical\_discussion ($-38.14$), neutral\_formal ($-29.60$), fear ($-16.83$) \\
        \hline
    \end{tabularx}
\end{table}
 
Trained on the full set of lexicon categories, the logistic regression classifier reached an accuracy of 0.7382 and a ROC-AUC of 0.8210. The technical\_discussion coefficient ($-38.14$) dwarfed everything else in magnitude, and sexual\_pleasure ($+19.37$) came out as the single strongest positive predictor, well clear of every other category. Neither result does much to ease our original concern, that these categories are picking up dataset- or source-specific artifacts rather than anything that generalizes, which is exactly why Model 2 exists and why it's the model we treat as primary for the rest of this study.
 
\begin{table}[htbp]
    \centering
    \caption{Model 2 (Leakage-Resistant Feature Set), Standardized: Main Result}
    \label{tab:model_2_standardized}
    \vspace{0.5em}
    \renewcommand{\arraystretch}{1.3}
    \begin{tabularx}{\linewidth}{| l | c | c |}
        \hline
        \rowcolor{tableheader}
        \textbf{Feature} & \textbf{Std. Coef.} & \textbf{Odds Ratio} \\
        \hline
        trust & $+0.701$ & $2.02$ \\
        \hline
        greed & $+0.466$ & $1.59$ \\
        \hline
        excitement & $+0.343$ & $1.41$ \\
        \hline
        urgency & $+0.228$ & $1.26$ \\
        \hline
        curiosity & $+0.104$ & $1.11$ \\
        \hline
        social\_connection & $+0.069$ & $1.07$ \\
        \hline
        information\_transactional & $+0.013$ & $1.01$ \\
        \hline
        request\_action & $+0.002$ & $1.00$ \\
        \hline
        authority & $-0.084$ & $0.92$ \\
        \hline
        fear & $-0.500$ & $0.61$ \\
        \hline
        neutral\_formal & $-0.958$ & $0.38$ \\
        \hline
    \end{tabularx}
    \caption*{\footnotesize Accuracy = 0.7082, ROC-AUC = 0.7397. Coefficients express the change in log-odds per one standard deviation increase in the feature; odds ratios are $e^{\beta}$.}
\end{table}
 
\begin{table}[htbp]
    \centering
    \caption{Model 2, Unstandardized (Reference)}
    \label{tab:model_2_performance}
    \vspace{0.5em}
    \renewcommand{\arraystretch}{1.3}
    \begin{tabularx}{\linewidth}{| l | X |}
        \hline
        \rowcolor{tableheader}
        \textbf{Metric} & \textbf{Value} \\
        \hline
        Accuracy & 0.6860 \\
        \hline
        ROC-AUC & 0.7595 \\
        \hline
        Top malicious predictors & greed (+20.65), trust (+20.64), excitement (+14.67), urgency (+12.58) \\
        \hline
        Top legitimate predictors & neutral\_formal ($-28.82$), fear ($-19.91$), authority ($-4.14$) \\
        \hline
    \end{tabularx}
\end{table}
 
After removing the three leakage-susceptible categories (technical\_discussion, sexual\_pleasure, promotion\_marketing), the model was retrained using the remaining features. The standardized Model 2 reached an accuracy of 0.7082 and a ROC-AUC of 0.7397. The unstandardized version shifted this balance, with a lower accuracy of 0.6860 and a slightly higher ROC-AUC of 0.7595; this is simply what happens when regularization is applied unevenly across features on different scales rather than any sign of a fundamentally different model underneath.

Table~\ref{tab:lexicon_perf} adds precision, recall, and F1 for the malicious class, and Table~\ref{tab:lexicon_cm} gives the confusion matrices. The errors are mostly false positives: the unstandardized Model 2 catches 7{,}392 of 8{,}569 malicious test emails but also flags 3{,}978 of 7{,}847 legitimate ones, so the weakness of the lexicon model is false alarms rather than missed attacks.

\begin{table}[htbp]
\centering
\caption{Lexicon models on the test split (malicious = positive class).}
\label{tab:lexicon_perf}
\footnotesize
\renewcommand{\arraystretch}{1.2}
\setlength{\tabcolsep}{4pt}
\begin{tabular}{|l|c|c|c|c|c|}
\hline
\textbf{Model} & \textbf{Acc.} & \textbf{AUC} & \textbf{Prec.} & \textbf{Rec.} & \textbf{F1} \\
\hline
Model 1 & 0.7382 & 0.8210 & 0.6996 & 0.8736 & 0.7770 \\
\hline
Model 2 & 0.6860 & 0.7595 & 0.6501 & 0.8626 & 0.7415 \\
\hline
Model 2 (std.) & 0.7082 & 0.7397 & 0.6712 & 0.8643 & 0.7556 \\
\hline
\end{tabular}
\end{table}

\begin{table}[htbp]
\centering
\caption{Confusion matrices on the test split (counts).}
\label{tab:lexicon_cm}
\footnotesize
\renewcommand{\arraystretch}{1.2}
\begin{tabular}{|l|c|c|c|c|}
\hline
\textbf{Model} & \textbf{TN} & \textbf{FP} & \textbf{FN} & \textbf{TP} \\
\hline
Model 1 & 4{,}632 & 3{,}215 & 1{,}083 & 7{,}486 \\
\hline
Model 2 & 3{,}869 & 3{,}978 & 1{,}177 & 7{,}392 \\
\hline
Model 2 (std.) & 4{,}219 & 3{,}628 & 1{,}163 & 7{,}406 \\
\hline
\end{tabular}
\end{table}
 
We read these results as modest but statistically grounded evidence that psychological and intent-based lexicon features offer an assistive signal rather than a standalone one. The lexicon-based models trail well behind the raw-text classifiers in Table~\ref{tab:model_performance}, which isn't surprising given the gap in feature space: our lexicon models compress each email down to about a dozen interpretable psychological dimensions, while the baseline classifiers work with thousands of lexical dimensions. That's not really a weakness to be explained away so much as a difference in purpose. The lexicon approach was never going to out-predict a high-capacity classifier on raw accuracy, and that was never really the point; its value is in being able to say \textit{why} a message reads as manipulative, which is something Table~\ref{tab:model_performance}'s classifiers can't offer on their own.
 
\subsection{Cross-Validation and Held-Out Source Results}
\label{sec:heldout}

Table~\ref{tab:cv_lexicon} reports 5-fold cross-validation for the lexicon models on the full deduplicated set. Table~\ref{tab:loso} reports the leave-one-source-out results; ROC-AUC is undefined for Nazario and Nigerian\_Fraud because every email in them is malicious, so only accuracy is given. The CV scores agree with the test split (standardized Model 2: ROC-AUC $0.7466 \pm 0.0056$ against 0.7397), so the single split was not unusually favorable. The leave-one-source-out results are mixed. Model 2's ROC-AUC ranges from 0.62 (CEAS\_08) to 0.84 (Ling), and its mean over the four sources containing both classes (0.71) is close to the 0.73 obtained from a random split of the same cleaned data. Model 1 also averages 0.71, so removing the three categories did not improve transfer on average, and Model 1 transfers better to CEAS\_08 (0.80 against 0.62). Accuracy at the default threshold is poor on some sources (0.29 on Nazario), so the features transfer as a ranking but the decision threshold does not. We therefore read these results as evidence of some cross-source signal, not as confirmation that the three removed categories were the main source of leakage.

\begin{table*}[t]
\centering
\caption{5-fold cross-validation of the lexicon models (mean $\pm$ sd).}
\label{tab:cv_lexicon}
\footnotesize
\renewcommand{\arraystretch}{1.2}
\begin{tabular}{|l|c|c|c|c|c|}
\hline
\textbf{Model} & \textbf{Acc.} & \textbf{AUC} & \textbf{Prec.} & \textbf{Rec.} & \textbf{F1} \\
\hline
Model 1 & 0.7448 $\pm$ 0.0027 & 0.8256 $\pm$ 0.0045 & 0.7066 $\pm$ 0.0027 & 0.8741 $\pm$ 0.0042 & 0.7815 $\pm$ 0.0022 \\
\hline
Model 2 & 0.6873 $\pm$ 0.0039 & 0.7601 $\pm$ 0.0033 & 0.6518 $\pm$ 0.0024 & 0.8609 $\pm$ 0.0066 & 0.7419 $\pm$ 0.0038 \\
\hline
Model 2 (std.) & 0.7080 $\pm$ 0.0042 & 0.7466 $\pm$ 0.0056 & 0.6737 $\pm$ 0.0025 & 0.8546 $\pm$ 0.0121 & 0.7534 $\pm$ 0.0051 \\
\hline
\end{tabular}
\end{table*}

\begin{table}[htbp]
\centering
\caption{Leave-one-source-out results for the lexicon models.}
\label{tab:loso}
\footnotesize
\renewcommand{\arraystretch}{1.2}
\setlength{\tabcolsep}{4pt}
\begin{tabular}{|l|c|c|c|c|}
\hline
\textbf{Held-out source} & \textbf{n} & \textbf{M1 AUC} & \textbf{M2 AUC} & \textbf{M2 Acc.} \\
\hline
CEAS\_08 & 32{,}670 & 0.8047 & 0.6216 & 0.5644 \\
\hline
Enron & 28{,}307 & 0.6172 & 0.6991 & 0.6269 \\
\hline
Ling & 2{,}854 & 0.7539 & 0.8432 & 0.7687 \\
\hline
SpamAssassin & 5{,}649 & 0.6792 & 0.6782 & 0.6479 \\
\hline
Nazario & 1{,}523 & -- & -- & 0.2869 \\
\hline
Nigerian\_Fraud & 3{,}229 & -- & -- & 0.7733 \\
\hline
\end{tabular}
\end{table}

\subsection{Mann-Whitney U Tests and Results}
\label{sec:mann-whitney}
 
\begin{table}[H]
    \centering
    \caption{Features Significantly Higher in Malicious Emails (Mann-Whitney U Test)}
    \label{tab:malicious_features}
    \vspace{0.5em}
    \renewcommand{\arraystretch}{1.3}
    \footnotesize\setlength{\tabcolsep}{3pt}
    \resizebox{\linewidth}{!}{\begin{tabular}{| l | c | c | c | c | c |}
        \hline
        \rowcolor{tableheader}
        \textbf{Feature} & \textbf{Mean (Mal.)} & \textbf{Mean (Legit.)} & \textbf{Effect Size} & \textbf{p-value} & \textbf{Cliff's $\delta$} \\
        \hline
        trust & 0.002757 & 0.000513 & $+0.002245$ & $\approx 0$ & $+0.151$ \\
        \hline
        greed & 0.003456 & 0.000928 & $+0.002528$ & $\approx 0$ & $+0.101$ \\
        \hline
        excitement & 0.002362 & 0.000745 & $+0.001617$ & $1.56\times10^{-143}$ & $+0.064$ \\
        \hline
        urgency & 0.003837 & 0.002243 & $+0.001595$ & $8.70\times10^{-17}$ & $+0.027$ \\
        \hline
        curiosity & 0.000797 & 0.000508 & $+0.000289$ & $2.13\times10^{-13}$ & $+0.013$ \\
        \hline
        social\_connection & 0.001406 & 0.001014 & $+0.000392$ & $4.57\times10^{-5}$ & $-0.010$ \\
        \hline
    \end{tabular}}
\end{table}
 
\vspace{1em}
 
\begin{table}[H]
    \centering
    \caption{Features Significantly Higher in Legitimate Emails (Mann-Whitney U Test)}
    \label{tab:legitimate_features}
    \vspace{0.5em}
    \renewcommand{\arraystretch}{1.3}
    \footnotesize\setlength{\tabcolsep}{3pt}
    \resizebox{\linewidth}{!}{\begin{tabular}{| l | c | c | c | c | c |}
        \hline
        \rowcolor{tableheader}
        \textbf{Feature} & \textbf{Mean (Mal.)} & \textbf{Mean (Legit.)} & \textbf{Effect Size} & \textbf{p-value} & \textbf{Cliff's $\delta$} \\
        \hline
        neutral\_formal & 0.002555 & 0.009732 & $-0.007177$ & $\approx 0$ & $-0.279$ \\
        \hline
        fear & 0.001070 & 0.003535 & $-0.002465$ & $\approx 0$ & $-0.145$ \\
        \hline
        request\_action & 0.009798 & 0.011704 & $-0.001906$ & $7.04\times10^{-117}$ & $-0.095$ \\
        \hline
        authority & 0.003136 & 0.004000 & $-0.000864$ & $3.17\times10^{-92}$ & $-0.069$ \\
        \hline
        information\_transactional & 0.004871 & 0.005831 & $-0.000960$ & $9.51\times10^{-215}$ & $-0.114$ \\
        \hline
    \end{tabular}}
\end{table}
 
All eleven leakage-resistant features came back statistically significant ($p < 0.0001$ across the board). That's not much of a surprise at $n = 65{,}662$ training emails, and it shouldn't be mistaken for evidence of practical importance on its own; what we actually care about is the direction and size of each effect. On that front, the strongest mean differences (neutral\_formal, greed, fear, and trust) are the same four features that carry the largest standardized coefficients in Model 2, which gives the model's learned weights some extra statistical footing beyond the coefficients by themselves. Cliff's delta, reported in the same tables, ranks the features similarly: neutral\_formal ($-0.279$), trust ($+0.151$), and fear ($-0.145$) have the largest magnitudes, while urgency ($+0.027$), curiosity ($+0.013$), and social\_connection ($-0.010$) are close to zero. The sign for social\_connection even disagrees with its mean difference, so we treat it as essentially no effect.

Fear and authority scored higher in legitimate emails, which at first runs against the usual social engineering account. We see three likely reasons, none of which we tested directly. First, our fear and authority lexicons are made of everyday workplace words (``risk,'' ``problem,'' and ``error'' for fear; ``department,'' ``team,'' and ``support'' for authority) that occur routinely in ordinary corporate and technical email. Second, the malicious emails here are largely spam and advance-fee fraud, which lean on rewards (greed, excitement, trust) rather than threats or impersonated authority, so fear- and authority-based phishing such as account-suspension notices is probably underrepresented. Third, the legitimate class includes corporate (Enron) email, where talking about problems and departments is normal. These words may therefore mark organizational text more than manipulation, and a per-source breakdown would show whether they also act as source markers.
 
\subsection{Special Case: The Urgency Lexicon Category}
\label{sec:urgency-case}
 
Urgency deserves its own discussion. In the unstandardized Model 2, it carries one of the largest positive coefficients of any category ($+12.58$), but raw coefficients depend on each feature's scale. On the standardized scale its coefficient is moderate ($+0.228$, odds ratio $1.26$), well below trust ($+0.701$) and greed ($+0.466$), which matches its fairly modest standalone effect size ($+0.001595$, against more than $+0.0022$ for trust and greed; Cliff's $\delta = +0.027$). Standardizing therefore removes the apparent mismatch between the coefficient and the effect size.
 
This episode is a decent illustration of why we keep coming back to effect size rather than p-values when judging how much a feature actually matters (Section~\ref{sec:mann-whitney}). At this sample size, almost everything clears the bar for statistical significance, so a tiny p-value doesn't tell us much about how large or meaningful an effect really is. Urgency's modest standalone effect also tracks with something fairly intuitive: urgency language shows up plenty in ordinary, legitimate corporate email too, a deadline reminder being the obvious case, which puts a natural ceiling on how much it can tell you by itself.

\subsection{Connecting Classification and Psychological Analysis}
\label{sec:connecting}

The two halves of this study answer different questions about the same problem. The raw-text classifiers in Table~\ref{tab:model_performance} ask whether an email is malicious, and they answer it very well, with accuracy between 0.98 and 0.99. The lexicon-based models in Table~\ref{tab:model_2_standardized} ask what about an email reads as manipulative, and they answer it with far less predictive power (ROC-AUC of 0.7397 for the standardized Model 2). It would be easy to read that gap as the lexicon approach simply losing, but the two sets of models are not really competing. The baseline classifiers draw on thousands of lexical dimensions, while the lexicon models compress each email into about a dozen interpretable ones, so the difference in performance largely reflects a difference in purpose.

That difference in purpose is also why the first part cannot replace the second. A classifier that reports 0.98 accuracy still hands an analyst, or an employee who just received a suspicious message, a verdict with no explanation attached. The lexicon scores supply the missing piece. An email flagged as malicious that also scores high on trust and greed looks like a reward-based lure built on manufactured credibility, which is a different situation from one that leans on urgency and authority. In a corporate setting, that distinction matters for triage and for security awareness training, and it is something the baseline models cannot offer on their own.

We want to be clear about what this study does and does not show. We did not test whether adding the lexicon features to the raw-text classifiers improves detection, and with the baselines already performing near ceiling, we would expect the headroom for any such improvement to be small. What the results do support is a complementary arrangement: a high-accuracy classifier flags incoming email in real time, and the psychological profile from the lexicon model is attached to explain what the message is attempting. Evaluating that combined pipeline directly, both for detection performance and for how useful the explanations are to human analysts, is a natural next step.
 
\subsection{Summary of Findings}
 
A few things stand out once all of this is put together. Emotion- and intent-based lexicon features carry a real, statistically grounded signal for detecting social engineering, and that signal holds up even after the dataset-specific features are stripped out (the standardized Model 2's ROC-AUC landed at 0.7397). We'd rather be precise about how that signal gets framed, though: it's assistive, not standalone. The lexicon-based models trail well behind the raw-text classifiers in Table~\ref{tab:model_performance}, and where they actually earn their keep is interpretability, not competing on raw accuracy. The content of the signal matched what the theory would predict, too: malicious emails leaned on trust-building, greed, and excitement and, more weakly, on urgency and curiosity, several of which map directly onto Cialdini's compliance principles \cite{cialdini2021_influence}, while legitimate emails leaned toward neutral formality, clear action requests, and an authoritative tone. The urgency case, on its own, is a useful reminder that raw coefficients can overstate a feature's role when features sit on different scales, and that standardized coefficients and effect size, not statistical significance alone, are the numbers worth trusting once sample sizes get this large.

\section{Limitations}
\label{sec:limitations}

Our results also depend on the dataset. All experiments use one collection of public emails; the leave-one-source-out test checks transfer between its component sources but not to a fully separate dataset, so we cannot say how well the findings hold on other data. Only part of this collection is corporate email, and much of the rest is spam or general fraud, which is not the same as targeted attacks like the manager impersonation case that motivated this work. Many of the sources are also old, and attack styles have changed since then. We tried to reduce dataset-specific effects by removing technical\_discussion, sexual\_pleasure, and promotion\_marketing, but we picked them by looking at the size of their coefficients, which is a rough method. Other features, such as neutral\_formal, may still reflect differences between sources and not real differences in intent.

The lexicon models also have modest performance. The standardized Model 2 reached an accuracy of 0.7082 and a ROC-AUC of 0.7397, and it misclassified a fair number of legitimate emails, so it would raise too many false alarms if used on its own. Logistic regression treats each feature as having a separate linear effect, and we did not include interaction terms, so any interaction between cues remains untested. In addition, the leakage-prone categories were chosen using data that included the held-out sources, so the leave-one-source-out results are not fully independent of that choice.

Finally, the scope of the study is limited. We only looked at English email text and did not use attachments, links, sender information, or other metadata. We did not test whether adding the lexicon features to the raw-text classifiers improves detection, and we did not ask security analysts whether the psychological profiles are useful to them in practice. Attackers could also change their wording once they know which language patterns are being checked. These points are left for future work.

\section{Conclusion}

The goal we aimed for with this study was to investigate how, if even, psychological and linguistic characteristics of email communications provide meaningful signals for detecting social engineering attacks. To achieve this, we used a publicly available combined email dataset (82{,}078 deduplicated emails) and developed a classification model using features derived from the content of the messages. Rather than merely distinguishing between malicious and benign emails, we looked at psychological and intent-related characteristics within the messages to better understand the persuasive strategies that may be associated with social engineering.

Our study shows that psychological manipulation can be seen through identifiable emotional and intent-based patterns in email text. In particular, malicious emails were more strongly associated with features such as urgency, excitement, trust, and curiosity, which may correspond to attempts to influence the recipient's emotional state and encourage a desired response. On the other hand, legitimate emails showed stronger associations with features such as neutral formality, requests for action, and authority. These findings suggest that social engineering detection can benefit from considering not only the surface-level content of an email, but also the psychological characteristics part of it.

To confirm whether these differences actually followed meaningful patterns rather than being random variations, we applied the Mann-Whitney U test to the extracted features. We also studied the possibility that certain features were capturing dataset-specific patterns rather than genuine characteristics of social engineering. After removing potentially problematic feature categories, the resulting leakage-resistant feature set continued to provide predictive value, achieving a ROC-AUC of 0.7397 on the test split. This shows us that psychological and intent-based features contain useful information for classifying malicious and legitimate emails, although tests on held-out email sources gave mixed results.

Overall, the findings demonstrate that psychological and intent-based analysis can be used to aid conventional approaches to social engineering detection. Rather than treating an email solely as malicious or benign, taking the emotion-based patterns within the message into account may provide useful insight into the mechanisms used by social engineering attacks. However, the results also show that these features should be considered as assisting signals rather than a standalone solution. Additionally, further evaluation on larger and more diverse datasets is necessary to establish whether this approach is generalizable.

Future work could extend this research beyond email to other forms of professional communication, including platforms such as Microsoft Teams and Slack. Additional datasets could also be incorporated to evaluate whether the observed psychological patterns remain consistent across communication platforms and organizational contexts. Furthermore, future studies could investigate more sophisticated language models and explainable detection methods to determine whether psychological manipulation can be identified reliably while maintaining interpretability for security analysts and end users.

\section*{Acknowledgment}

The study was supported by the School of Computing Sciences and Computer Engineering at the University of Southern Mississippi. We also acknowledge the assistance of our lab mentor, Mr. Oluseyi Olukola, throughout the project.

\bibliographystyle{IEEEtran}
\bibliography{References}

\end{document}